# DESIGN, IMPLEMENTATION, AND COOPERATIVE COEVOLUTION OF AN AUTONOMOUS/TELEOPERATED CONTROL SYSTEM FOR A SERPENTINE ROBOTIC MANIPULATOR

Don Sofge & Gerald Chiang
GreyPilgrim Inc.
687-J Lofstrand Lane
Rockville, MD 20850
Email: dsofge@greypilgrim.com
Ph: (301) 610-6393x207

ABSTRACT

This paper describes the design, implementation, and machine learning issues associated with developing a control system for a serpentine robotic manipulator. The purpose of the control system is to provide autonomous/teleoperative control of the serpentine robotic manipulator, as well as full robotic control during operation of the manipulator within an enclosed environment such as an underground storage tank.

The controller algorithms make use of both low-level joint angle control employing force/position feedback constraints, and high-level coordinated control of end-effector positioning. Since the inverse kinematics solutions for a hyper-redundant serpentine manipulator are extremely difficult if not impossible to obtain using standard techniques, a variety of methods from evolutionary computation were employed to arrive at a suitable inverse kinematics model for the manipulator. Additional constraints, such as various joint angle bending restrictions, joint constraints within a given stage of the manipulator, and customer requirements may be easily incorporated into the machine learning process through the fitness function.

This approach has resulted in a system which offers both high-level full robotic control and low-level telerobotic control modes, and provides a high level of dexterity for the operator.

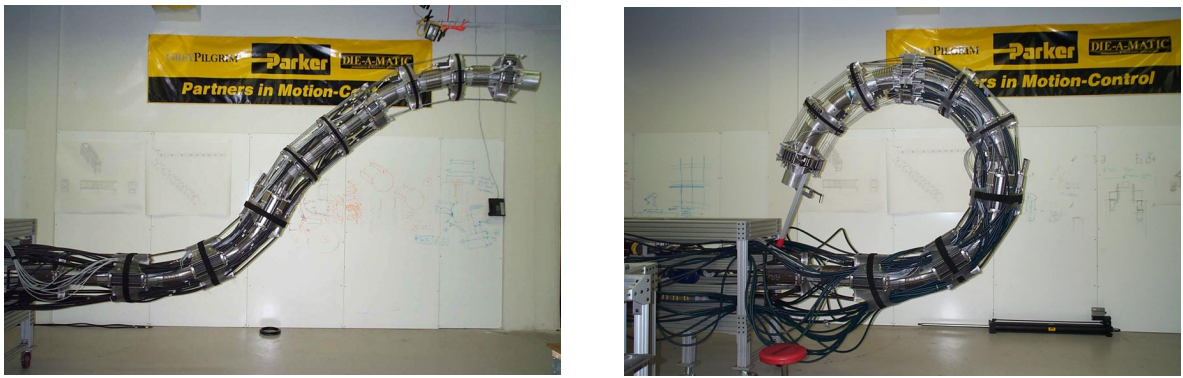

Figure 1. GreyPilgrim's "EMMA" Hyper-Redundant Robotic Manipulator

# 1. INTRODUCTION: HIGH-LEVEL CONTROL OF HYPER-REDUNDANT MANIPULATORS

Traditional robotic manipulators typically consist of a small number of links, each sequential pair connected by a joint. However, another class of robots is based upon the design concept of a large number of segments, with each segment pair separated by a joint or flexible member of some sort. These robots (see Figure 1) have been variously described as snake-like manipulators[1], tentacles, elephant trunks[2], continuum robots[3], high degree of freedom (HDOF) robots, or hyper-redundant manipulators[4]. In this work these robotic designs will be referred to as hyper-redundant manipulators, since this term is used most frequently in the references to the literature on analysis of the kinematics (both forward and inverse) of such robotic systems.

Hyper-redundant manipulators derive their name from the fact that they have many redundant degrees of freedom. In traditional robotic manipulator design, one needs only 6 degrees of freedom (DOFs) to achieve any position and orientation within the robot's workspace. Three degrees are used for spatial positioning (x, y, z) and three are used for orientation (roll, pitch, yaw). Redundancy is defined in terms of the manipulator's ability to achieve a given position or orientation in more than one way. A hyper-redundant manipulator consisting of a series of alternating rigid segments and flexible joints has very many possible joint configurations (in fact, infinitely many) which result in the same end-effector position.

# 2. PREVIOUS WORK ON HYPER-REDUNDANT ROBOTS

## 2.1 Historical Perspective

Much of the work in designing and building hyper-redundant manipulators was pioneered by Hirose[1] of Japan. Hirose was attempting to understand the mechanisms behind locomotion of snakes and snake-like creatures, so he studied the gaits of various snakes, built a variety of snake-like robots, and attempted to recreate those gaits in his snake-like robots.

The difficulty in solving for the inverse kinematics of a hyper-redundant manipulator is generally not in finding a solution to achieve a desired end-effector position, but rather in finding a set of solutions with constraints on the redundant degrees of freedom such that one can move the end-effector smoothly through a desired trajectory, and that the resulting joint motion will not be excessive (ideally, this motion will be minimized).

A rigorous mathematical analysis of inverse kinematics for hyper-redundant manipulators was performed by Chirikjian[5]. Chirikjian used techniques from differential geometry to describe robotic kinematics as fitting curves in space, and proposed a novel type of hyper-redundant robot known as a variable geometry truss (VGT) structure. The differential geometry approach describes a curve in space as a series of moving frames originating at the base of the manipulator, progressing along the length and ending at the end-effector. The curves used are taken from a basis set which tend to form S-shaped curves. Once a set of curves is determined to approximate a desired configuration, then the manipulator is fitted to the curve set.

Chirikjian[6] and others[7] have attempted to use the Frenet-Serret formulation for curves in space, but have found it to be too computationally expensive to use in real-time for calculating inverse kinematics. Related advanced techniques by Gravagne[3] use other formulations for these curves (such as wavelet functions), but require many simplifying assumptions about the manipulator design (such as equal segment lengths, or homogeneous bend of the backbone material), and these techniques have only been applied to 2-D (planar) robots. Before the work discussed herein, no general approach has been presented for finding the inverse kinematics of hyper-redundant manipulators in a three-dimensional setting.

The work discussed in this paper uses techniques from machine learning to constrain the extra degrees of freedom for hyper-redundant manipulators, and find a solution set of configurations such that smooth robotic motions are generated both for joint angles and end-effector motion.

2.2 Machine Learning Approach

Evolutionary computational techniques provide a variety of powerful methods for solving optimization problems. If the problem of finding the inverse kinematics for a hyper-redundant manipulator is appropriately cast as an optimization problem, or series of optimization problems, then we can apply EC techniques. In order to do this we need to specify the "cost" or "fitness" function(s) in terms such that they reflect how well (or poorly) a particular solution fits as part of the desired inverse kinematics solution set.

The first step is to discretize the workspace of the robot into a grid of regularly spaced points. For each point, we can then use an evolutionary strategy (with a population size of one and a mutation operator), and then from a random legal starting configuration (set of joint angles for the robot) evolve the configuration until the end-effector is within a desired error tolerance of the goal point in the grid. The cost function (which is to be minimized) is simply the distance in 3-D Euclidean space from the end-effector to the goal point. After each goal point has been reached, the corresponding set of joint angles (configuration of the robot) is stored in a look-up table. Note that in working through the grid, it is not necessary to restart at a random configuration each time. After a target has been reached, it is quite convenient to move to a neighboring grid point. However, this may introduce a bias into the configuration set dependent upon the predominant direction of motion through the grid.

After a configuration is found for each grid point within the discretized workspace, then the configuration set may be co-evolved so that for neighboring grid points, the distance in joint space between each configuration and its neighbors (that is, the two-norm of the difference between the configuration vectors) is minimized. In order to do this, a second constraint is added to the cost function (or alternatively a second cost function may be specified). The first constraint, we recall, is the distance to target. The second constraint is a penalty function defined as the sum of the distances in joint space between each configuration and its neighbors in the grid. For example, with a 2-D grid space

$$\text{penalty}_{i,j} = \sum\sum |(C_{i,j} - C_{k,m})| \quad \text{summed over } k=(i-1,i+1), m=(j-1,j+1) \quad (1)$$

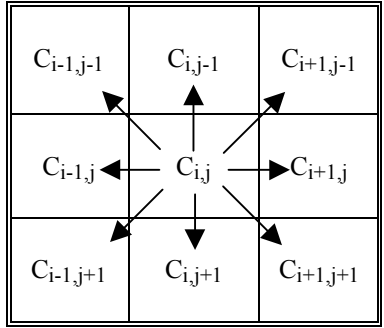

Figure 2. Penalty Function for 2-D Grid Space

Once we have discretized the robotic workspace into sufficiently small cells, with each cell having four corners as the grid points, we can co-evolve the configuration set such that each configuration is within the desired error tolerance of having the end-effector hit the target, and that the penalty for each configuration (based upon differences between itself and its neighbors) is minimized. The configurations are then stored in a look-up table. We can then use the stored look-up table to provide the set of joint angles to position the end-effector at any grid point within the workspace.

What then do we do if our desired target location doesn't fall on a grid point whose configuration (set of joint angles) is stored in our lookup table? If the penalty terms are so well minimized that the configurations are close enough together in joint space, then we can simply perform a bi-linear interpolation (assuming a 2-D grid) based upon the configurations of the grid points that are vertices of the cell in the workspace where our desired location falls (if we have a 3-D grid, then the linear interpolation is simply performed in three-space using 8 grid points, the corners of a cube). Therefore, with a sufficiently well behaved solution space, we can smoothly move the end-effector of the robotic manipulator through a continuous trajectory, and the deviation of the end-effector from each target point along the trajectory will be quite small. (Another way of looking at this is that the inverse kinematics is stored as a look-up table, and local linear approximation is used to retrieve the data. If the level of resolution of the look-up table is high enough, any continuous nonlinear function can be approximated arbitrarily closely by local linear interpolation of the look-up table values).

## 3. SIMULATION RESULTS

As a means for proving this methodology, a 3-D simulation of a serpentine hyper-redundant robotic manipulator was created. In the simulated system the arm consists of a 13-ft mast, 5 manipulator stages, a wrist and a gripper end-effector. The 13-ft. mast is attached to the actuation package (which is at the top), followed by 3 rigid segments approximately 11-ft. in length each (stages 1-3), stage 4 which is composed of 3 rigid segments separated by an equal number of flexible couplings, stage 5 which is also composed of 3 rigid segments separated by flexible couplings, and a wrist and end-effector assembly. The mast is capable of rotation +/-180 degrees, stages 1-3 have hinge joints that bend in only the pitch direction 0-45 degrees. Stage 4 couplings may each bend in both pitch and yaw directions, but while the pitch range for each coupling is +/-30 degrees, the yaw range is +/-15 degrees. Stage 5 couplings may pitch or yaw +/-30 degrees in each direction. In both stages 4 and 5, the bend of the joints is constrained by the fact that each segment in these stages is not individually actuated, but actuated as the entire stage is bent. The constraint used in the simulation was that all of the pitch angles in stage 4 were the same, as were the yaw angles. Similar constraints were assumed for stage 5. Thus, an assumption of equal bend throughout the stage (for 4 & 5) for each independent degree of freedom was made. The wrist also has two degrees of freedom (roll and pitch), +/-180 degrees for roll, +/-60 degrees for pitch. A tank enclosure was also modeled in the simulation (see Figure 3 below).

As shown in Figure 3, the manipulator was simulated inside the tank enclosure. The simulation model included full forward kinematics of the manipulator, so that given a set of joint angles (20 in all), the simulated arm can be correctly positioned, and the location of the end-effector can be precisely determined.

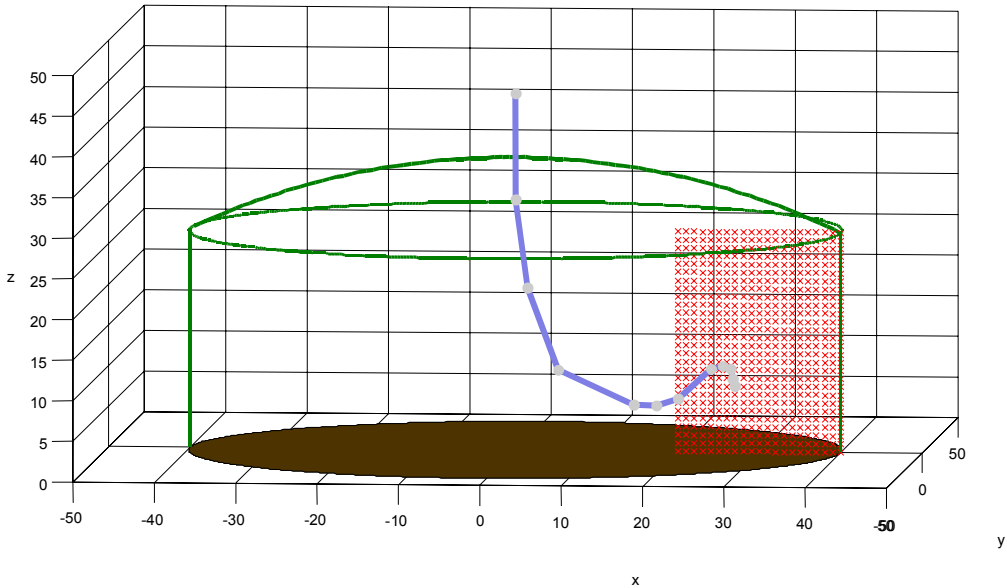

Figure 3. 3-D Graphical Simulation of Serpentine Manipulator within a Tank

A grid of points (shown as small points on the figure) which represents a 21x28 ft. cross-section of the tank was defined such that the points in the grid are separated by one foot intervals in the horizontal and vertical directions. Taking advantage of the symmetry of the task, this plane of grid points may be rotated through the tank (by rotation of the mast), so that if we have a solution for this plane, we can use it to position the end-effector anywhere that can be expressed as a combination of an (x, z) coordinate pair on the grid plane, plus a rotation θ about the z-axis. A similar technique was used to map out a grid of points under the dome above the side-wall level of the tank.

An evolutionary strategy was implemented to optimize the joint configuration for each point in the grid. The ES algorithm used only a population of one and only a mutation operator. However, since the fitness (or distance to target) was highly coupled in the 20 individual joint angles, a form of search was necessary that was more directed than random mutation of the entire vector. Each of the joint angles in the configuration was sequentially mutated (starting at the base), with the provision that a positive mutation (resulting in higher fitness) was kept, while a negative mutation was discarded. An exponentially decaying annealing factor was maintained for determining the variance of the mutation (the mean was zero). The "count" used in the decay was based upon the number of positive mutations generated. Because each joint angle required adjustments of varying magnitudes in order to move the end-effector toward the target, a separate annealing factor was maintained for each vector component. This proved to be a very effective strategy, once an appropriate initial magnitude of the variance was chosen (e.g. 10 degrees). Getting tight convergence to within less than one inch of the target for all configurations required manually adjusting the initial magnitudes and the decay rates (usually 0.5-0.9999), during learning. The system did converge within perhaps a few hundreds of passes through the data (the grid points) and periodic adjustment of annealing rates and initial variance of the mutation operator.

Given an inverse kinematic configuration to achieve each target point within the tank to within one inch error (a somewhat arbitrarily chosen tolerance), a penalty function was implemented as described above and co-evolution of the configuration space was attempted. The first method tried was to co-evolve the configurations by reducing the penalty while not allowing the distance to target for each configuration to suffer. This turned out to be ineffective. Only after the distance constraint was relaxed was the system able to co-evolve the configurations to reduce the penalty.

In addition, a third technique was used which required that each configuration continue to evolve to improve both its distance fitness (that is, get the distance to target as close to zero as possible), for each positive mutation it was required to reduce the penalty as well. This worked fairly well and effectively spread the general shapes of the configurations out across the grid. However, as might be expected some configurations eventually got "stuck" in the sense that a minimum had been reached so they could no longer improve. Due to this effect, it was necessary to implement a series of resets for reinitializing the variances of the mutation operators, counting the numbers of failed attempts to generate positive mutations and resets, and to eventually "give up" on a point until the next pass.

Progress could be made by alternately co-evolving the configurations to reduce penalties and distances, but not both at the same time. However, by using this method and adjusting the learning factors (starting variance of mutation operator and annealing rate), it was not difficult to get the configuration set to converge such that all of the solutions had very low penalties (resulting in smooth joint-space transitions across neighborhoods) and were within the one-inch distance tolerance.

The final configuration set stored in look-up table format was used to develop a controller for the simulated arm in the tank. As described previously, the 2-D plane was used to represent a full 3-D sweep through the tank (by decomposing any valid desired end-effector position into an [x, z] pair and rotation angle of the entire arm).  Continuous indexing into the look-up table was implemented by performing bi-linear interpolation of the cell vertex configurations for any desired configuration within a cell.  It was found that that smooth and accurate trajectory motion is achievable using the look-up table inverse kinematics method provided that the configurations are well co-evolved such that penalties are minimized and small, as defined with respect to some smoothness criterion (e.g. limiting joint velocities to a fixed maximum rate).

4. CONCLUSIONS

This work shows that inverse kinematics for hyper-redundant manipulators in a three-dimensional setting may be determined using evolutionary computation techniques.  Additional constraints, such as joint angle bending restrictions, joint constraints within a given stage of the manipulator, and customer requirements such as "stay off the hard stops and keep all parts of the manipulator no lower than the end-effector", may be easily incorporated into the learning process through the fitness function.  This approach is applicable for learning the inverse kinematics for a wide variety of designs of hyper-redundant robots, and does not require simplifying assumptions or 2-D planar manipulators as much of the related work in this area does.  In fact, at present this method is the only practically implementable general purpose method for determining the inverse kinematics in three-dimensions for multi-linked hyper-redundant manipulators.  While the inverse kinematic solution set obtained using this approach is not unique, it does possess an observable smoothness of motion that results from the co-evolution of configurations.  This smoothness of motion allows for smooth trajectory generation and following (using the end-effector) from any point to any other point within the workspace.